%% file: main.tex
\documentclass[10pt,twocolumn,letterpaper]{article}

 \usepackage[pagenumbers]{cvpr} 

\input{preamble}

\definecolor{cvprblue}{rgb}{0.21,0.49,0.74}
\usepackage[pagebackref,breaklinks,colorlinks,citecolor=cvprblue]{hyperref}

\def\paperID{199} 
\def\confName{3DV\xspace}
\def\confYear{2025\xspace}
\title{Space-time 2D Gaussian Splatting for Accurate Surface Reconstruction \\under Complex Dynamic Scenes}

\author{Shuo Wang$^{1}$\quad
Binbin Huang$^{2}$ \quad
Ruoyu Wang$^{1}$ \quad
Shenghua Gao$^{2}$\thanks{Corresponding author} \\
\\
$^{1}$ShanghaiTech University \quad
$^{2}$The University of Hong Kong \quad
\\
{\tt\small   \{wansghuo2022, wangry3\}@shanghaitech.edu.cn}\\{\tt\small binbinhuang@connect.hku.hk}, {\tt\small gaosh@hku.hk}\\
}
\begin{document}
\maketitle
\input{sec/0_abstract}    
\input{sec/1_intro}

\input{sec/2_related}

\input{sec/3_method}
\input{sec/4_experiment}
\input{sec/5_conclusions}
{\small\bibliographystyle{ieeenat_fullname}
\bibliography{main}
}
\input{sec/X_suppl}
\end{document}

%% file: preamble.tex
\usepackage[table,dvipsnames]{xcolor}%

\usepackage{multirow}
\definecolor{tabfirst}{rgb}{1, 0.7, 0.7} 
\definecolor{tabsecond}{rgb}{1, 0.85, 0.7} 
\definecolor{tabthird}{rgb}{1, 1, 0.7} 
\usepackage{pifont}
\usepackage{booktabs}
\usepackage{colortbl}
\usepackage{graphicx}
\usepackage{subcaption}
\usepackage{xspace}
\usepackage{multirow}
\usepackage{makecell}
\usepackage{algorithm}

\newcommand{\bc}{\mathbf{c}}\newcommand{\bC}{\mathbf{C}}

\newcommand{\bM}{\mathbf{M}}
\newcommand{\bn}{\mathbf{n}}\newcommand{\bN}{\mathbf{N}}

\newcommand{\bp}{\mathbf{p}}
\newcommand{\bq}{\mathbf{q}}
\newcommand{\br}{\mathbf{r}}
\newcommand{\bS}{\mathbf{S}}
\newcommand{\bt}{\mathbf{t}}

\newcommand{\bx}{\mathbf{x}}

\newcommand{\cG}{\mathcal{G}}

\newcommand{\cS}{\mathcal{S}}

\makeatletter
\DeclareRobustCommand\onedot{\futurelet\@let@token\@onedot}
\def\@onedot{\ifx\@let@token.\else.\null\fi\xspace}
 
\def\ie{i.e\onedot}

\makeatother

\definecolor{yellow}{rgb}{1, 1, 0.7}
\definecolor{orange}{rgb}{1, 0.85, 0.7}
\definecolor{tablered}{rgb}{1, 0.7, 0.7}
\definecolor{red}{rgb}{1, 0, 0}

\definecolor{wincolor}{rgb}{0.85, 0.0, 0.0}

\definecolor{darkyellow}{rgb}{0.8, 0.8, 0.5}
\definecolor{darkred}{rgb}{0.7, 0.3, 0.3}
\definecolor{darkgreen}{rgb}{0.3, 0.7, 0.3}
\definecolor{green}{rgb}{0, 1.0, 0}
\definecolor{pink}{rgb}{1, 0.4, 0.7}



%% file: sec/0_abstract.tex
\begin{abstract}
 Previous surface reconstruction methods either suffer from low geometric accuracy or lengthy training times when dealing with real-world complex dynamic scenes involving multi-person activities, and human-object interactions. 
 To tackle the dynamic contents and the occlusions in complex scenes, we present a space-time 2D Gaussian Splatting approach. Specifically, to improve geometric quality in dynamic scenes, we learn canonical 2D Gaussian splats and deform these 2D Gaussian splats while enforcing the disks of the Gaussian located on the surface of the objects by introducing depth and normal regularizers. Further, to tackle the occlusion issues in complex scenes, we introduce a compositional opacity deformation strategy, which further reduces the surface recovery of those occluded areas. Experiments on real-world sparse-view video datasets and monocular dynamic datasets demonstrate that our reconstructions outperform state-of-the-art methods, especially for the surface of the details. The project page and
more visualizations can be found at: \url{https://tb2-sy.github.io/st-2dgs/}. 

 
\end{abstract}

%% file: sec/1_intro.tex
\section{Introduction}
\label{sec:intro}
\begin{figure}[t]	\includegraphics[width=1.\linewidth]{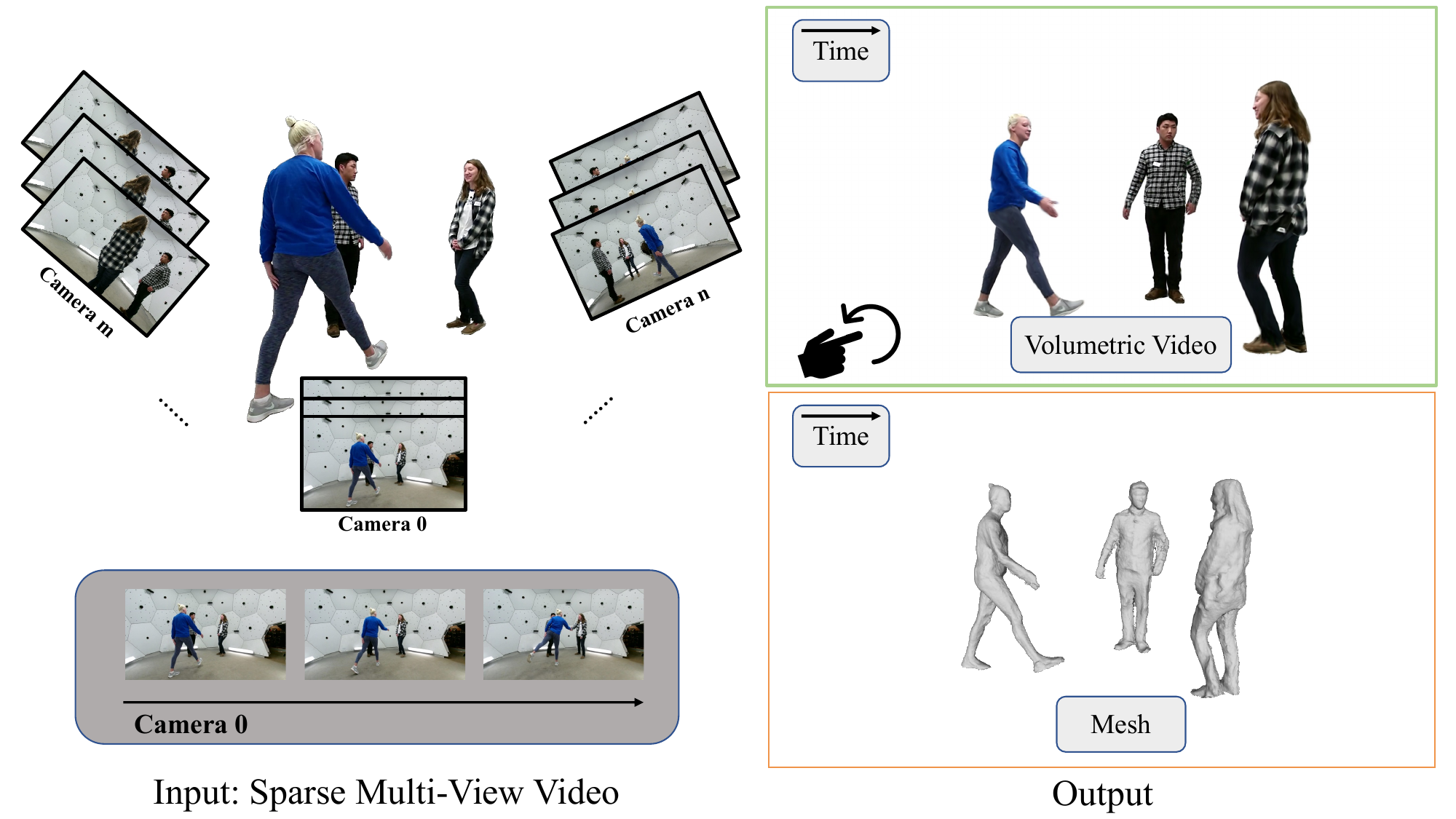}
 \caption{Given sparse view or monocular video input, our method achieves high-quality surface reconstruction and rendering of dynamic scenes.}
 \vspace{-1em}
\label{fig:teaser}
\end{figure}

Capturing accurate geometry in complex dynamic scenes from sparse-view videos (see Figure~\ref{fig:teaser}) remains a significant challenge. These scenes often involve severe occlusions, and the dynamic nature of the content requires the surface reconstruction to adapt consistently over time, further complicating the task.

Traditional approaches rely on depth data from RGBD sensors to build mesh models~\cite{dynamicfusion, surfelwarp}, but these methods are prone to holes, noise, and limited texture detail in the depth maps. With the rise of neural rendering, it has become possible to generate 4D neural surfaces with photo-realistic textures using sparse multi-view cameras~\cite{sdfflow, shao2023tensor4d}, monocular videos~\cite{ndr, tinuevox, kplanes}, advanced dome systems~\cite{joo2015panoptic}, and even in uncontrolled environments~\cite{li2023dynibar, dynamicnerf, nsff}. However, many of these techniques use volumetric density fields~\cite{tinuevox, kplanes} or signed distance functions~\cite{ndr, shao2023tensor4d, sdfflow} to represent surfaces, which often results in high storage requirements, long training times, and potential reductions in rendering quality. While more compact approaches, such as hash grids~\cite{ngp, neus2} and octrees~\cite{plenoctrees, fourieroctrees}, have been developed, they primarily focus on reconstructing single objects. 

The limitations of purely volumetric representations have spurred renewed interest in point-based graphics~\cite{dynamicfusion,surfelwarp,dynamicpointfield}, which offer advantages in scalability, efficiency, and persistent surface tracking—critical benefits for dynamic reconstruction. Early point-based pioneering work~\cite{aliev2020neural, rakhimov2022npbg++, xu2022point, dynamicpointfield} uses points, enhanced with neural networks, as rendering primitives for both static and dynamic scenes. Although ~\cite{dynamicpointfield} also targets animated meshes, it often relies on predefined topology or accurate surface normals, which are challenging to acquire in practice. The advancement of point-based representation is exemplified by 3D Gaussian Splatting, which excels in appearance synthesis and achieves unprecedented rendering speeds, maintaining interactive frame rates. Motivated by the successes of 3D Gaussian Splatting, many subsequent methods have extended point-based representations to 4D scene representation~\cite{dynamic3dgs, gs4d, deformable3dgs, real4dgs, 4dslice, Li_STG_2024_CVPR}, but almost all of them focus on novel view synthesis rather than geometry reconstruction. 

To address surface reconstruction, some studies have focused on animating 3D Gaussians based on template meshes, primarily for specific applications like avatars or human performance capture~\cite{gsavatr, 3dgsavatar}. Another approach, Vidu4D~\cite{vidu4d}, targets dynamic scene reconstruction from monocular videos. It optimizes time-varying warping functions for transforming Gaussian surfels, resulting in high-quality geometry and realistic rendering of dynamic objects. However, its application has been limited to simpler objects or slower-motion scenarios.

To address the challenge of accurate geometry recovery, Huang \emph{et al.}~\cite{2dgs} introduced a surfel-based 2D Gaussian Splatting (2DGS) method that places Gaussian disks on object surfaces. While effective for static objects and novel view synthesis, 2DGS is not designed for dynamic scenes. To overcome this limitation, we propose a novel space-time 2D Gaussian Splatting approach. Our method learns a set of canonical Gaussian splats, and for each timestamp, a deformation network adjusts these splats to capture dynamic scene changes. By jointly optimizing the canonical 2DGS with the rendered images at each timestamp, we ensure the deformed Gaussian disks adhere closely to object surfaces. This provides a highly accurate surface representation with straightforward geometry regularization, utilizing depth and surface normals, which are essential for reconstructing dynamic surfaces.

We introduce a time-varying opacity model that improves geometric precision and minimizes noise to tackle the occlusion challenges common in sparse-view or multi-object interactions. Additionally, to remove textureless backgrounds, we apply a regularizer that aligns the rendered alpha mask with the foreground mask, speeding up the learning process. Our approach produces a geometrically accurate surface representation that can efficiently convert into a triangle mesh using TSDF fusion~\cite{curless1996volumetric}.

Our key contributions are as follows: \begin{itemize} \item We present space-time 2D Gaussian Splatting, the first particle-based surface model for complex dynamic scene reconstruction, achieving faster processing than traditional volumetric SDF approaches while maintaining high geometric accuracy. \item We introduce a joint optimization of canonical and deformed 2DGS across different timestamps, enabling precise surface reconstruction in dynamic scenes. Additionally, we propose a time-varying opacity model to address occlusion and improve geometric fidelity. \item We demonstrate the effectiveness of our method on challenging dynamic datasets, including D-NeRF~\cite{dnerf} and CMU Panoptic~\cite{cmu}. Our approach surpasses state-of-the-art methods in both quality and efficiency. \end{itemize}

%% file: sec/2_related.tex
\section{Related Work}
\label{sec:related}

\subsection{Dynamic Scenes Novel View Synthesis}
Dynamic scenes involve moving objects and changing lighting conditions, necessitating models that can accommodate temporal variations and motion. Some pioneering works have addressed these challenges by incorporating temporal dimensions into NeRF frameworks~\cite{park2021hypernerf, dnerf, park2021nerfies}. Subsequently, other methods focus on improving efficiency~\cite{cao2023hexplane, kplanes}, self-supervised decoupling~\cite{d2nerf, Chen_2023_CVPR, Cheniccv2023}, investigating explicit representations~\cite{cages, gs4d, dynamic3dgs}, and advancing image-based rendering~\cite{li2023dynibar}. 3D Gaussian Splatting~\cite{3dgs} (3DGS) has gained popularity due to its high rendering quality and real-time rendering speed. However, when directly applied to dynamic scenes, it can result in blurring due to the lack of temporal modeling. Previous works~\cite{gs4d, deformable3dgs, real4dgs, 4dslice} have extended 3D Gaussian splatting to include temporal aspects, but there has been limited research on the geometric accuracy of dynamic scenes using Gaussian splatting. 2D Gaussian Splatting~\cite{2dgs} (2DGS) retains the advantages of high-quality and real-time rendering seen in 3D Gaussian Splatting, while addressing the issue of multi-view geometric inconsistency in 3D Gaussian splatting. Additionally, it offers new approaches for geometric modeling in dynamic scenes. We have taken advantage of the above advantages of 2DGS and established space-time 2DGS to conveniently model geometrically accurate dynamic scenes. 
\subsection{Surface reconstruction of dynamic scenes.}
Obtaining accurate meshes from a scene is a critical and long-standing task in computer vision, with many studies~\cite{wang2021neus, volsdf,sugar,2dgs} dedicated to this objective. Most previous work has focused on static scenes, achieving relatively accurate results. Some approaches use implicit function methods~\cite{wang2021neus,volsdf,monosdf}, while others~\cite{sugar,2dgs,surfels,gof,chen2024pgsr,zhang2024rade} employ explicit point-based methods. Certain methods~\cite{banmo,vid2avatar, gsavatr, zhao2024surfel} for geometric reconstruction of dynamic scenes often require assumptions such as the presence of articulated or templates to maintain the complete geometric shape. However, these assumptions are not generally applicable, and such methods frequently fail in real-world complex scenes.  Existing methods~\cite{ndr, shao2023tensor4d, sdfflow} can nonetheless handle unconstrained scenes. Tensor4D \cite{shao2023tensor4d} captures dynamic scenes using a 4D tensor representation, which is then decomposed into multiple 2D planes to improve training and inference efficiency. SDFFlow~\cite{sdfflow} no longer directly estimates the SDF value at each moment but instead outputs the derivative of the SDF value, integrating it from the previous moment to obtain the current SDF value. Although it achieves high geometric accuracy, this method is very time-consuming. DG-Mesh~\cite{dgmesh} and MaGS~\cite{mags} leverages 3D Gaussian splatting by matching triangles and Gaussians to improve geometric consistency. However, due to DG-Mesh's strict requirement of Gaussians anchoring to triangles, the Gaussian has more uniform scales. Many small Gaussians are discarded to enhance rendering quality in difficult areas and the method is only tested in simple single-object scenes. While our approach leverages Space-time 2D Gaussian primitives for dynamic surface approximation, it combines the rendering quality and accuracy geometry.

%% file: sec/3_method.tex
\begin{figure}
    \centering
    \includegraphics[width=\linewidth]{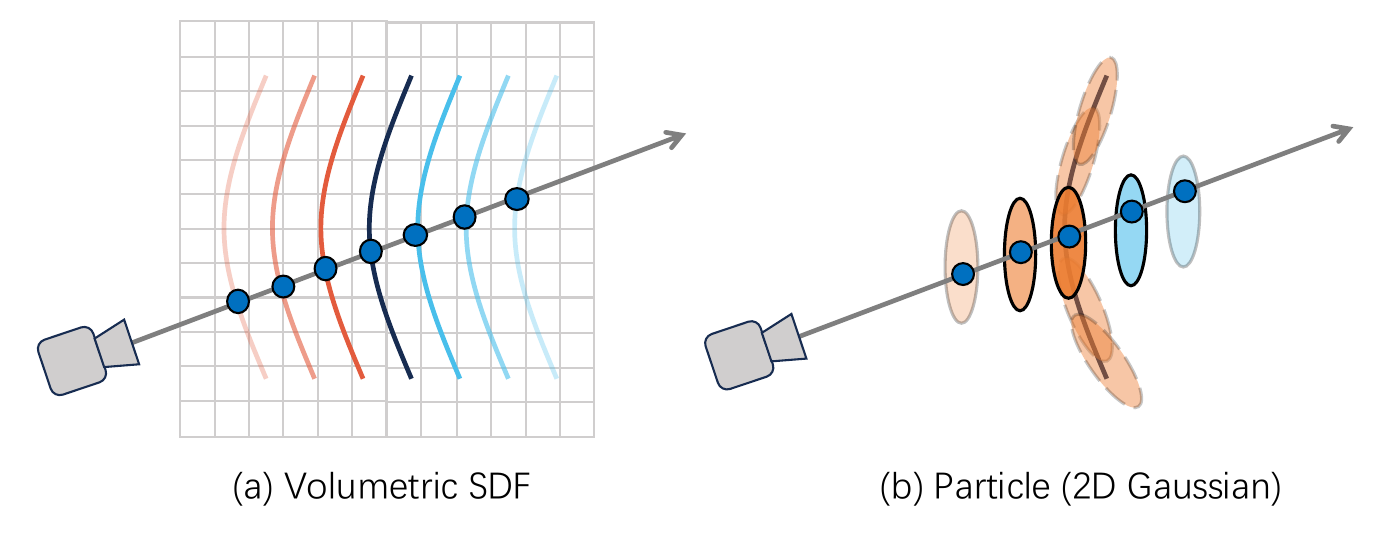}
    \caption{Two different scene representations. SDF-based methods~\cite{wang2021neus, volsdf} capture surfaces at specific locations, while our particle-based representation utilizes discrete 2D Gaussians~\cite{2dgs}. This particle-based method is more storage-efficient, offers faster rendering, and simplifies motion tracking.}
    \label{fig:comparison}
\end{figure}

\section{Preliminaries}
In contrast to traditional volumetric representations such as Signed Distance Functions (SDF), our approach employs 2D Gaussian splatting~\cite{2dgs} to represent scenes. This section revisits two common volumetric surface reconstruction methods, \ie, volume-based and particle-based representation, as illustrated in Figure~\ref{fig:comparison}. These methods are rendered using a unified volume rendering equation:

\begin{equation}
\bC(\br) = \sum_{i=1} \bc(\bx_i),\alpha(\bx_i)\ \prod_{j=1}^{i-1} (1 - \alpha(\bx_j))
\label{eq:volume}
\end{equation}
In this equation, $\bC(\br)$ represents the color of a ray $\br$, and $\bx_i$ are the sample points along the ray. Here, $\bc(\bx)$ denotes the color field, and $\alpha(\bx)$ represents the opacity, which is derived from continuous SDF fields~\cite{wang2021neus,volsdf} or density fields~\cite{nerf}.

Unlike NeRF-based surface reconstruction, which captures surfaces at specific locations $\bx$, our method uses 2D Gaussian splatting \cite{2dgs} to represent surfaces with discrete particles. Each particle is defined by a set of parameters: $\cG = \{\bp, \bt_u, \bt_v, s_u, s_v, o, \bc\}$, where $\bp$ is the center, $\bt_u$ and $\bt_v$ are the tangential vectors, $s_u$ and $s_v$ are the scalings factors, $o$ is the opacity, and $\bc$ is the color.
The volume rendering equation \eqref{eq:volume} is adapted to accumulate colors by intersecting rays with these particles. Specifically, the alpha value for a 2D Gaussian splat along a ray $\br$ is calculated as:
\begin{equation}
\alpha(\bx) = o\exp\left(-\frac{u(\bx)^2+v(\bx)^2}{2}\right) \label{eq:alpha}
\end{equation}
where $u(\bx)$ and $v(\bx)$ are local coordinates in the tangent space, computed as:
\begin{align}
u(\bx) = (\bx - \bp)^{\mathrm{T}} \bt_u / s_u  \\
v(\bx) = (\bx - \bp)^{\mathrm{T}} \bt_v / s_v
\end{align}

The compact particle representation of 2D Gaussian splatting offers advantages in terms of memory efficiency and rendering speed compared to direct volume rendering. This method is detailed further in the original paper~\cite{2dgs}. Additionally, the compact nature of this representation facilitates easier motion tracking. This approach can be categorized under Eulerian and Lagrangian representations from a simulation perspective. In the following sections, we extend this approach to dynamic scenes and address the associated challenges.

\section{Space-time 2D Gaussians}
To model space-time 2D Gaussians, we define a set of 2D Gaussians in a canonical space, denoted as $\cS = {\cG^c_i}$. For simplicity, we omit the index $i$. Each 2D Gaussian is represented by $\cG^c = \{\bp^c, \bt_u^c, \bt_v^c, s_u^c, s_v^c, o^c, \bc^c\}$. Following the method described in~\cite{2dgs}, tangent vectors are modeled as quaternions $\bq_c$, and scalings are represented by a matrix $\bS$. In the following, we parameterize $\cG^c = \{\bp^c, \bq^c, \bS^c, o^c, \bc^c\}$ as a set of \textit{pre-activated} tensors. 

\paragraph{Geometry Deformation.} To animate these 2D Gaussians within the canonical space, we introduce a deformation field $\Phi(\bx, t)$, which takes a location $\bx$ and a time-step $t$ as inputs and produces offsets that deform the Gaussian's position and shape. Specifically:
\begin{align}
(\Delta\bp, \Delta\bq, \Delta\bS) = \Phi(\bx, t) \\
(\bp^t, \bq^t, \bS^t) = (\bp^c + \Delta\bp, \bq^c + \Delta\bq, \bS^c + \Delta\bS)
\end{align}
We implement the deformation field as a Hex-Plane, similar to 4DGS~\cite{gs4d}, for its efficiency. However, the deformation can be realized with any 4D function architecture. The resulting parameters are then activated and rasterized by the differential surfel rasterizer\footnote{https://github.com/hbb1/diff-surfel-rasterization}~\cite{2dgs} to render images from specific viewpoints.

\begin{figure*}[t]
\includegraphics[width=1.\linewidth]{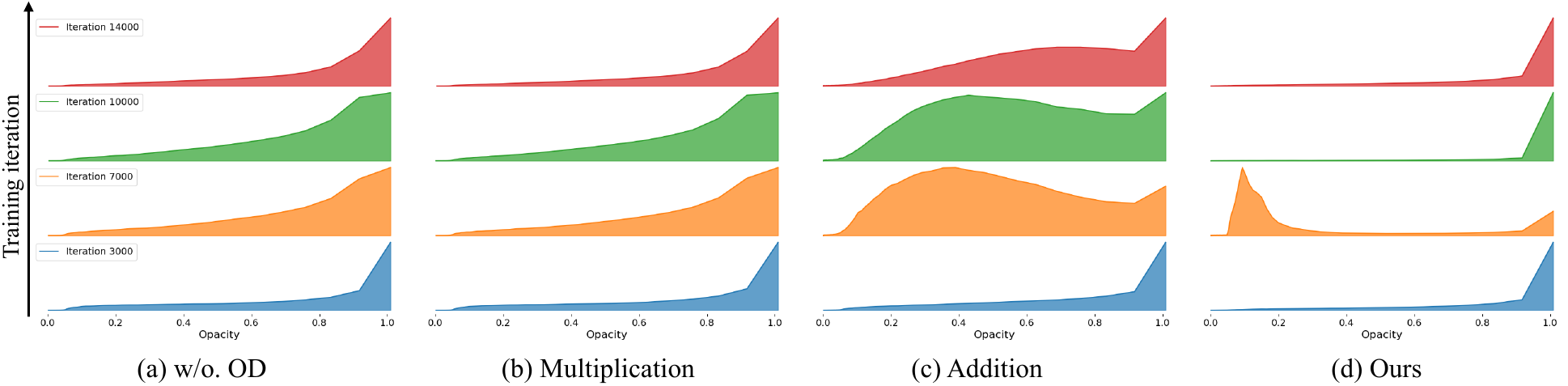}
 \caption{
Different opacity deformation approach for \texttt{Pizza1} scene where canonical space opacity distribution (smooth results) over training iterations. a) Without opacity deformation and b) Multiplication~\cite{zhao2024gaussianprediction}, the opacity distribution gradually decreases from the larger end, failing to maintain a binary opacity state. c) Additon~\cite{gs4d} results in an unstable distribution. d) Our approach
allows canonical 2D Gaussians to maintain a stable and clear binary opacity state.}
 \vspace{-1em}
	\label{fig:train_opacity}
\end{figure*}

\paragraph{Opacity Deformation.} The 2D Gaussian representation includes an opacity parameter $o$ that models its visibility. A key consideration is how to handle opacity changes over time for dynamic scene reconstruction. According to the equation \eqref{eq:alpha}, the alpha value $\alpha(\bx)$, which governs the visibility of the Gaussian, is determined by both its opacity and shape. Excessive opacity changes can disrupt the learning of shape parameters and the geometry deformation field, potentially leading to unrealistic shape alterations, as observed by~\cite{gs4d}. On the other hand, neglecting to model opacity changes can impede the accurate representation of sudden movements, such as the appearance of flames.
\begin{figure}[t]	\includegraphics[width=1.\linewidth]{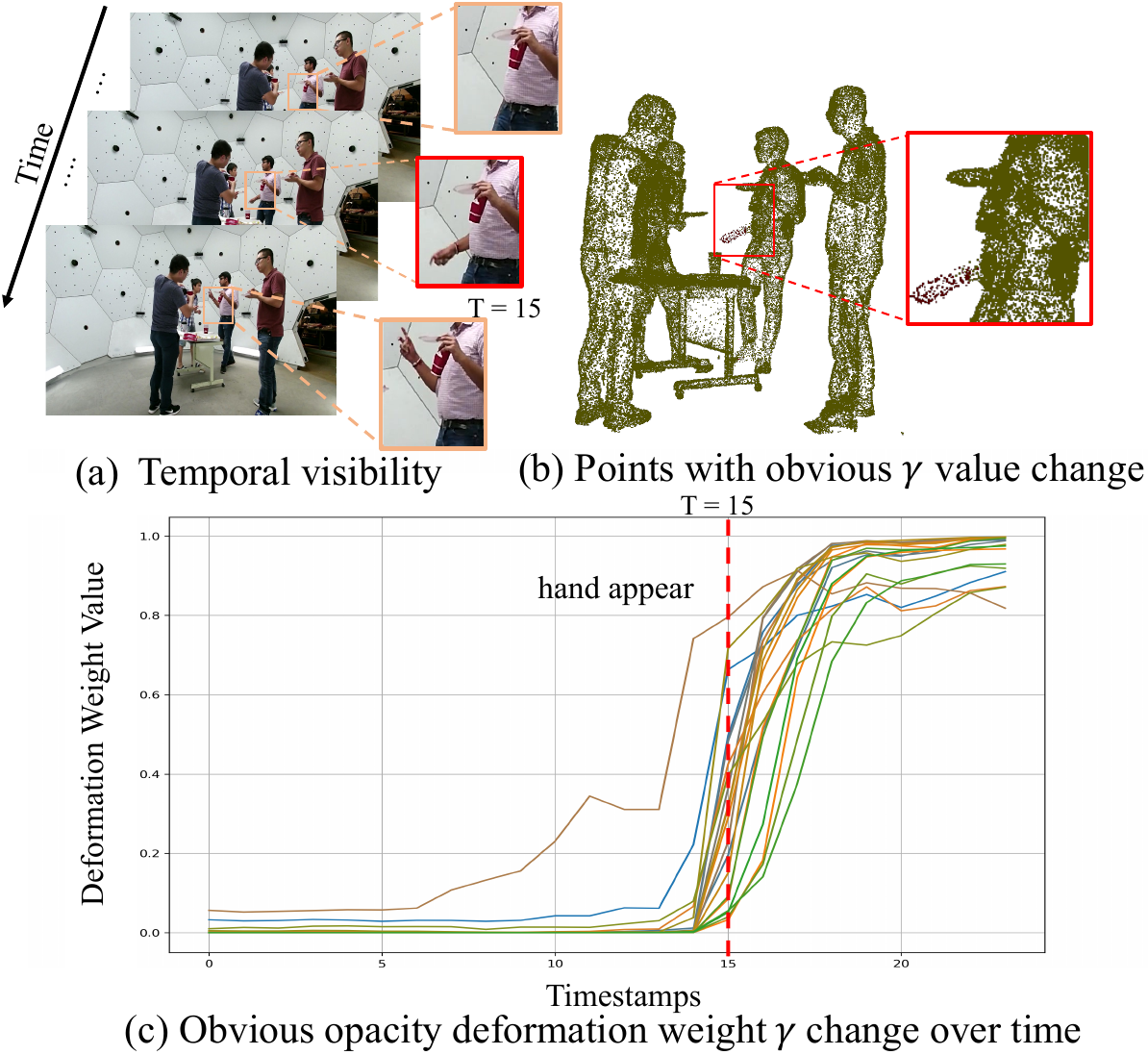}
 \caption{
 In (a), we illustrate the elements of sudden appearance. In (c), we visualize the changes in opacity deformation weight $\gamma$ over time. The points with significant changes in opacity weight are marked in red in (b), which corresponds to the sudden appearance of elements shown in (a). This further demonstrates the effectiveness of our proposed opacity deformation.
 }
 \vspace{-1em}
 \label{fig:visibility}
\end{figure}

In dynamic surface reconstruction, particularly in complex scenes with severe occlusion, certain regions may suddenly become visible, resembling phenomena such as ``spewing flames", yet these areas should remain solid. For example, as shown in Figure~\ref{fig:visibility}, a hand that was not visible in earlier frames appears suddenly at $T=15$. Without accounting for deformable opacity that adapts to changes in visibility, this sudden emergence would depend on distant, movable particles. This may lead to irrational shape and appearance. Therefore, accurately modeling opacity changes is essential for capturing this sudden appearance accurately.

However, integrating opacity deformation with space-time 2D Gaussians presents significant challenges, particularly in maintaining a well-distributed representation within canonical space. For example, if a 2D Gaussian $G^c$ is not accurately positioned, it may either move or have its opacity reduced to disappear.  Given that our scenes primarily consist of solid objects, enforcing a clear binary opacity state simplifies the training process.
We model opacity deformation $\Theta(\bx^c, t)$ similarly to geometry deformation. We compare three techniques for opacity deformation:
\begin{itemize}
\item \textbf{Addition}~\cite{gs4d}: $o^t = \varphi(\Theta(\bx_c, t) + o^c)$, where the deformed opacity can vary over time.
\item \textbf{Multiplication}~\cite{zhao2024gaussianprediction}: $o^t = \varphi(o^c) \cdot \gamma(\Theta(\bx_c, t))$ with $\gamma(x) \in (0,1)$, where the opacity decreases over time, inversely scaling $\varphi(o^c)$.
\item \textbf{Composition}~(Ours): $o^t = \varphi(o^c \cdot \gamma(\Theta(\bx_c, t)))$ with $\gamma(x) \in (0,1)$, where the deformed opacity $o^t$ are neutralized which inversely binarizes the $\varphi(o^c)$.
\end{itemize}

Both $\varphi$ and $\gamma$ are sigmoid functions. Figure~\ref{fig:train_opacity} illustrates the opacity distribution $\varphi(o^c)$ during training. Our method maintains a binary-opacity state, which aids in achieving a well-distributed representation of canonical 2D Gaussians. The addition of a bidirectional shift to $o^c$ leads to instability in optimization, as demonstrated in Figure~\ref{fig:train_opacity}\textcolor{red}{(c)}. Multiplication yields similar results to the case without opacity deformation, where $\gamma(\Theta(\bx_c, t))$ tends to approach 1. This is because $\gamma(\Theta(\bx_c, t)) \in (0,1)$ causes $o^t$ to be greater than or equal to $\varphi(o^c)$. When $\gamma$ is less than 1, the values with relatively low opacity in $o^c$ remain relatively high, which hinders optimization. Our method learns a stable binary opacity state by adversarially countering the neutralized inverse binarization of $\gamma(\Theta(\bx_c, t))$.

\paragraph{Optimization.} We begin by initializing the canonical 2D Gaussians in one of two ways: either by first performing a random initialization followed by fitting a canonical 2D Gaussian to all training frames in a monocular setting, or by directly performing a random initialization and then fitting a canonical 2D Gaussian to a set of posed images from a specific frame. Next, we use deformation fields to animate the shapes and opacities of these 2D Gaussians. During training, we dynamically add points, with hyper-parameters set according to 4DGS~\cite{gs4d}. The optimization process involves minimizing a composite loss function that includes photometric loss and additional regularization terms, as outlined in ~\cite{2dgs}.
The loss function we minimize is defined as:
\begin{equation}
\mathcal{L} = \mathcal{L}_{c} + \alpha \mathcal{L}_{d} + \beta \mathcal{L}_{n} + \eta \mathcal{L}_{m}
\end{equation}

Here, $\mathcal{L}_{c}$ represents the RGB reconstruction loss, combining $\mathcal{L}_1$ with the D-SSIM term from~\cite{kerbl3Dgaussians}. $\mathcal{L}_{d}$ and $\mathcal{L}_{n}$ are depth and normal regularization losses that ensure the 2D Gaussians form a coherent surface:
\begin{align}
 \mathcal{L}_{d} = \sum_{ij}w_i\|z_i-z_j\|_2 \\
 \mathcal{L}_{n} = \sum_{i}w_i(1-\|\bn_i^T\bN\|)
\end{align}
Here, $w_i=\alpha_i\prod_{j=1}^{i-1}(1-\alpha_j)$ is the $i$-th blending weight, $z_i$ is the $i$-th intersected depth and $\bn_i$ is the normal vector of 2D Gaussian, and $\bN$ is the pseudo normal vector derived from the rendered depth map. The regularization encourages volume rendering approaching surface rendering. Detailed explanations of these terms can be found in ~\cite{2dgs}.

In scenarios where only the foreground is modeled, these regularization terms can introduce background artifacts. To address this issue, we incorporate an additional alpha mask loss. We render the alpha mask as follows:
\begin{equation}
\bM(\br) = \sum_{i=1} \alpha(\bx_i) \prod_{j=1}^{i-1} (1 - \alpha(\bx_j)) \label{eq:mask_rendering
}
\end{equation}
and minimize it against a mask $\bM^*$ obtained following~\cite{sdfflow}, with the loss defined as:
\begin{equation}
\mathcal{L}_m = \sum_{\br}\|\bM(\br) - \bM^*(\br)\|
\label{eq:mask_loss
}
\end{equation}
Since masks may be less accurate at boundaries, we use this mask loss only in the early stages of training and reduce its influence by linearly decreasing $\eta$ as training progresses.
\begin{figure*}[htp!]	\includegraphics[width=1.\linewidth]{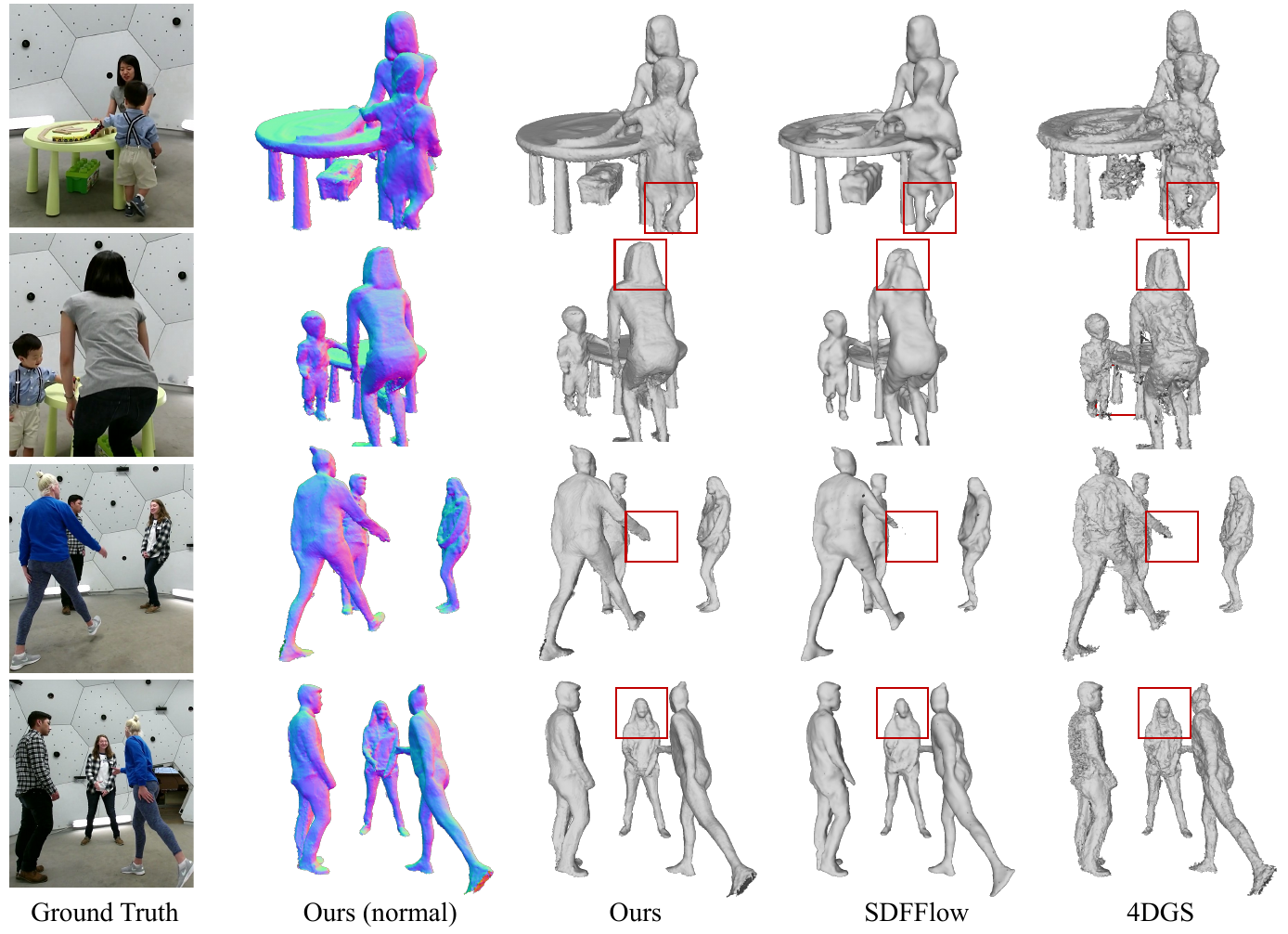}
 \caption{Visual comparisons at two different timestamps between our method, SDFFlow~\cite{sdfflow}, and 4DGS~\cite{gs4d} are conducted using scenes from a real-world dataset~\cite{cmu}. Our method, along with 4DGS, captures more details than the SDFFlow by leveraging the advantages of a point-based approach. However, because 4DGS is an extension of 3DGS, it suffers from insufficient geometric accuracy and produces a noisy surface.}
 \vspace{-1em}
\label{fig:main1}
\end{figure*}
\input{table/sdf_flow}

%% file: table/sdf_flow.tex
\begin{table}[t]
  \centering
  \resizebox{0.98\linewidth}{!}{
  \begin{tabular}{lcccccc}
    \toprule
    acc (mm) $\downarrow$& Ian3 & Haggling\_b2 & Band1 & Pizza1  & avg \\
    \midrule
     {NDR}~\cite{ndr}& 21.8 & 12.5 & 15.9 & 17.7  & 17.0 \\
     {Tensor4D~\cite{shao2023tensor4d}} & 15.4 & 13.7 & 17.1 & 18.3  & 16.1 \\
     {SDFFlow~\cite{sdfflow}} & 14.1 & \cellcolor{tabfirst}8.3 & 13.0 & \cellcolor{tabsecond}11.5  & 11.7 \\
     {4DGS} & \cellcolor{tabsecond}8.9 & 10.6 & \cellcolor{tabsecond}12.6 &  14.0  & \cellcolor{tabsecond}11.5 \\
      {Ours} & \cellcolor{tabfirst}8.6 & \cellcolor{tabsecond}8.6 & \cellcolor{tabfirst}11.9 & \cellcolor{tabfirst}11.2  & \cellcolor{tabfirst}10.1 \\
    \midrule
    comp (mm) $\downarrow$& Ian3 & Haggling\_b2 & Band1 & Pizza1  & avg \\
     \midrule
     {NDR~\cite{ndr}} & 20.7 & 22.8 & 23.7 & 25.0  & 23.1 \\
     {Tensor4D~\cite{shao2023tensor4d}} & 22.8 & 25.3 & 29.2  & 23.5 & 25.2 \\
     {SDFFlow~\cite{sdfflow}} & 17.5 & \cellcolor{tabfirst}18.6 & \cellcolor{tabsecond}21.4 & \cellcolor{tabsecond}20.6  & \cellcolor{tabsecond}19.5 \\
     {4DGS} & \cellcolor{tabsecond}16.8 &  19.2 &  21.6 &  22.0  &  19.9 \\
     {Ours} & \cellcolor{tabfirst}16.5 & \cellcolor{tabsecond}18.9 & \cellcolor{tabfirst}20.9 & \cellcolor{tabfirst}20.4  & \cellcolor{tabfirst}19.2 \\
    \midrule
    overall (mm) $\downarrow$& Ian3 & Haggling\_b2 & Band1 & Pizza1  & avg \\
    \midrule
     {NDR~\cite{ndr}} & 21.3 & 17.7 & 19.8 & 21.3  &  20.0 \\
     {Tensor4D~\cite{shao2023tensor4d}} & 19.1 & 19.5 & 23.2 & 22.9  & 21.2 \\
     {SDFFlow~\cite{sdfflow}} &  15.8 & \cellcolor{tabfirst}13.5 &  17.2 & \cellcolor{tabsecond}16.1  & \cellcolor{tabsecond}15.7 \\
     {4DGS} & \cellcolor{tabsecond}12.9  &  14.9 & \cellcolor{tabsecond}17.1 &  18.0  & \cellcolor{tabsecond}15.7 \\
      {Ours } & \cellcolor{tabfirst}12.6 & \cellcolor{tabsecond}13.7 & \cellcolor{tabfirst}16.4 & \cellcolor{tabfirst}15.8  & \cellcolor{tabfirst}14.6 \\
    \bottomrule
  \end{tabular}
  }
  \caption{Quantitative results on the CMU Panoptic dataset. 
  We report chamfer distance (acc, comp, overall) measured with all frames. The \colorbox{tabfirst}{best} and  \colorbox{tabsecond}{second} are marked with pink and orange. }
  \vspace{-5mm}
  \label{tab:cmu}
\end{table}

%% file: sec/4_experiment.tex
\section{Experiments}
\subsection{Implementations}
We implement our method using PyTorch~\cite{pytorch}
framework and do all experiments on a single NVIDIA A40 GPU. The optimization parameters of Gaussian points are the same as the original 3DGS~\cite{3dgs}. Following 4DGS~~\cite{gs4d}, we model the deformation field as a Hex-Plane and a tiny MLP decoder, with the initial learning rate of $1.6 \times 10^{-3}$, which is decayed to $1.6 \times 10^{-4}$ at the end of training. We set the number of initial and training iterations to $3K$ and $30K$, respectively, and we stop densifying and pruning Gaussian points at the iteration of $15K$. We adopt opacity cull and densify operation and set opacity cull thread to 0.05. The loss weight of $\mathcal{L}_d$, $\mathcal{L}_n$ and $\mathcal{L}_m$ are 1000, 0.05 and 1 for all scenes.  

\noindent\textbf{Mesh Extraction.} We follow the approach outlined in \cite{2dgs} and use Truncated Signed Distance Function (TSDF) fusion~\cite{curless1996volumetric} for mesh extraction. This method combines depth maps from all training views to fuse a holistic mesh. Since our approach uses only a limited number of views for training, it may result in aliasing artifacts in under-observed mesh regions. To address this issue, we interpolate pseudo views between the sparse training views, which helps to achieve smoother mesh extraction.

\subsection{Datasets and Evaluation Metrics}
\textbf{Datasets.}
We conduct quantitative experiments of geometry accuracy on the CMU Panoptic dataset~\cite{cmu}. We follow SDFFlow~\cite{sdfflow} to use the images from 10 RGB-D cameras positioned around the scene. The ground-truth point cloud at each timestamp is obtained by registering the depth maps taken from those cameras using the provided camera poses and intrinsic parameters. We use the 4 challenging complex dynamic scene clips: \texttt{Ian3}, \texttt{Haggling b2}, \texttt{Band1} and \texttt{Pizza1} for evaluation. 
Each clip contains 240 images, which are taken from the 10 cameras at 24 timestamps. The resolution of the images is 1920 × 1080. Given our objective of geometry accuracy, we utilize all 240 images for training and evaluate the reconstructed meshes.

We evaluate the quality of novel view synthesis on synthetic D-NeRF datasets~\cite{dnerf}. This dataset is designed for monocular settings with each scene containing only one randomly generated camera at every timestamp. The number of timestamps ranges from 50 to 200 between scenes. We select two classic scenes for experiments: \texttt{Lego} and \texttt{Mutant}. 
Besides, we also give a qualitative comparison of meshes since it lacks geometry ground truth.
\\

\textbf{Metrics.}
We assess our approach using the Chamfer distance, which includes metrics: accuracy, completeness, and overall distance. Considering the ground-truth point cloud $P$ and the predicted point cloud $\bar{P}$, the accuracy and completeness metrics are formulated as:
\begin{equation}
\text{Acc} = \frac{1}{|\bar{P}|} \sum_{\bar{p} \in \bar{P}} \min_{p \in P} \| p - \bar{p} \|_2
\end{equation}

\begin{equation}
\text{Comp} = \frac{1}{|P|} \sum_{p \in P} \min_{\bar{p} \in \bar{P}} \| p - \bar{p} \|_2
\end{equation}
The overall distance is then computed as the average of these two metrics. For our novel view synthesis experiments, We evaluate the results using peak-signal-to-noise ratio (PSNR),
perceptual quality measure LPIPS~\cite{lpips}, and the structural similarity index (SSIM)~\cite{ssim}.

\subsection{Performance Comparisons}

\begin{figure*}[h]	\includegraphics[width=.95\linewidth]{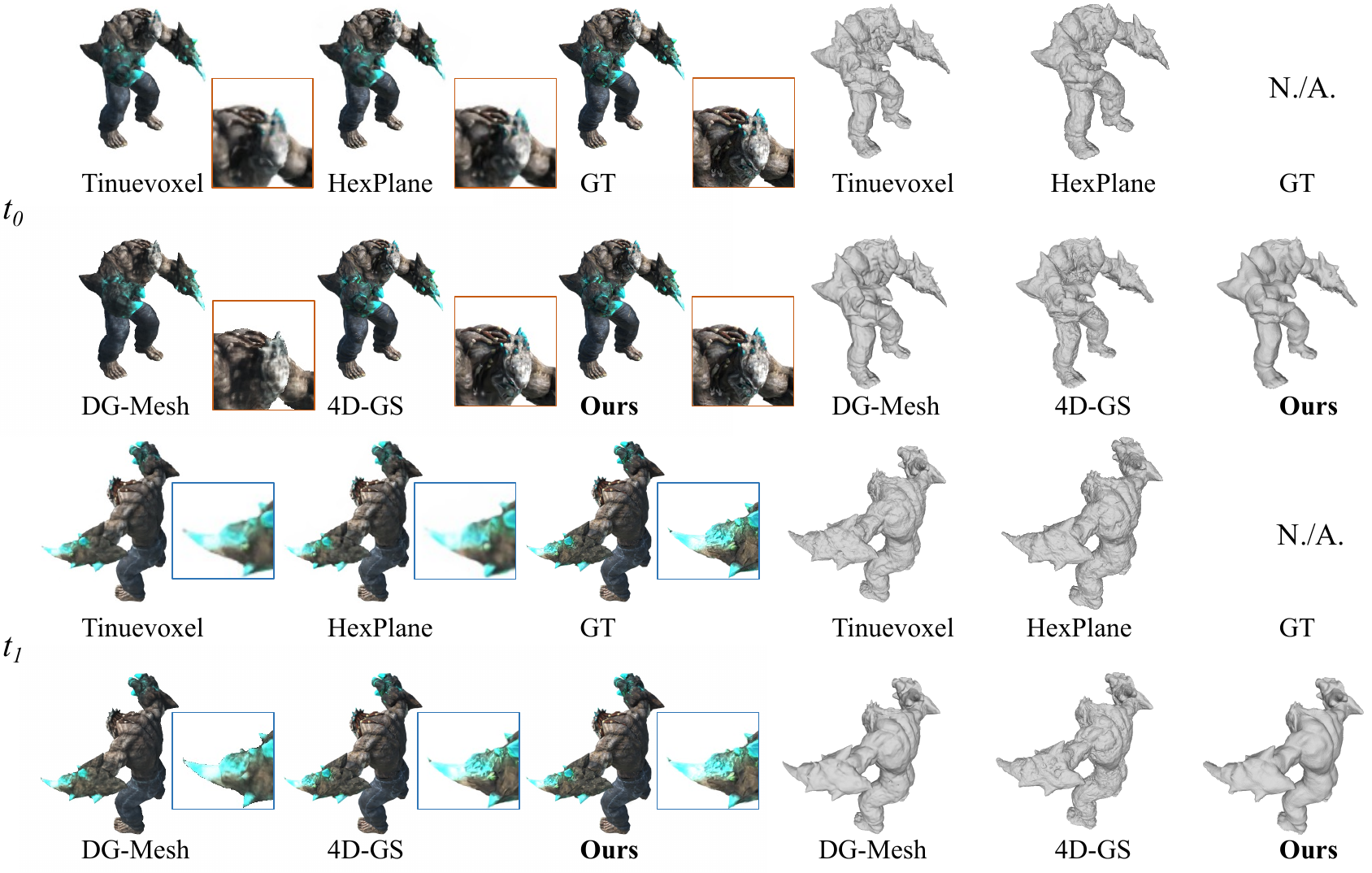}
 \caption{Qualitative comparison on the D-NeRF benchmark~\cite{dnerf}. Our method achieves comparable rendering quality compared with SOTA methods, producing more detailed and noise-free surfaces.}
	\label{fig:main2}
\vspace{-5mm}
\end{figure*}
\textbf{Comparisons on CMU Panoptic Dataset.} We compare our method against both implicit-based methods (SDFFlow~\cite{sdfflow}, Tensor4D~\cite{shao2023tensor4d} and NDR~\cite{ndr}) and explicit-based methods (4DGS~\cite{gs4d})
on the CMU Panoptic dataset. The quantitative results are shown in Table~\ref{tab:cmu}.
Our approach models the motion of Gaussian points explicitly and incorporates occlusion-adaptive attributes for Gaussian opacity, which achieves $12.1\%$ improvements in accuracy and $2\%$ improvements in completeness, thus making overall $7.5\%$ improvements compared with SDFFlow and 4DGS. Notably, benefiting from our explicit representation, our method can achieve efficient training ($100\times$ faster than SDFFlow, $10\times$ faster than Tensor4D) and real-time rendering as shown in Table~\ref{tab:training_time}. We also show qualitative comparisons in Figure~\ref{fig:main1}, which demonstrate that our method achieves more details and less noise.

\input{table/dnerf_res_sub}
\input{table/training_time}

\noindent\textbf{Comparisons on D-NeRF Dataset.}
Table~\ref{tab:main_res_dnerf} and Figure~\ref{fig:main2} present quantitative and qualitative comparisons among the previous SOTA methods and our approach on the D-NeRF dataset, which is designed for monocular settings. We benchmark our method with volume-based method ( TiNeuVox~\cite{tinuevox}, HexPlane~\cite{cao2023hexplane}), point-based methods (3DGS~\cite{3dgs}, 4DGS~\cite{real4dgs}) and a recent hybrid point-and-mesh representation (DG-Mesh~\cite{dgmesh}). Our approach achieves comparable rendering quality while offering higher reconstruction accuracy. Specifically, while TiNeuVox and HexPlane yield satisfactory results, they require longer rendering time. Point-based methods (3DGS, 4DGS) offer more efficiency but compromise on geometric accuracy. DG-Mesh achieves high-quality reconstruction but at the cost of lower rendering performance due to its reliance on the mesh-based renderer. Our method gets the best worlds, delivering fast speeds, high-quality rendering, and accurate reconstruction.

\begin{figure}[t]
\includegraphics[width=.98\linewidth]{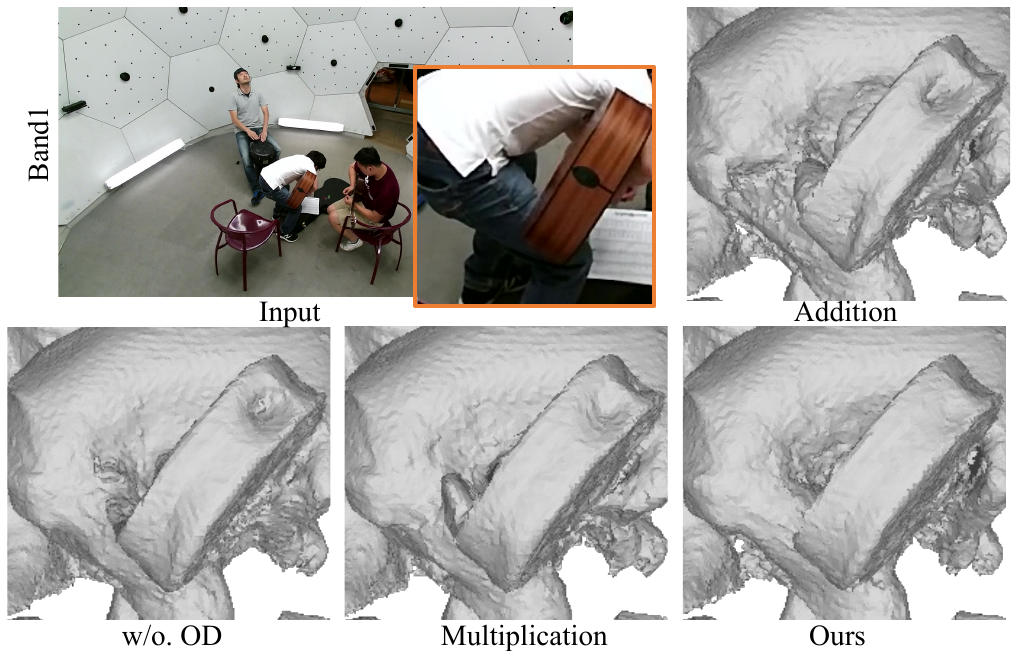}
 \caption{Quantitative ablation studies over different opacity deformation methods. Our method achieves a more rational shape.}
 \vspace*{-5pt}
 \label{fig:ablate_od}
\end{figure}

\subsection{Ablation Studies}
\begin{figure}[t]	\includegraphics[width=.98\linewidth]{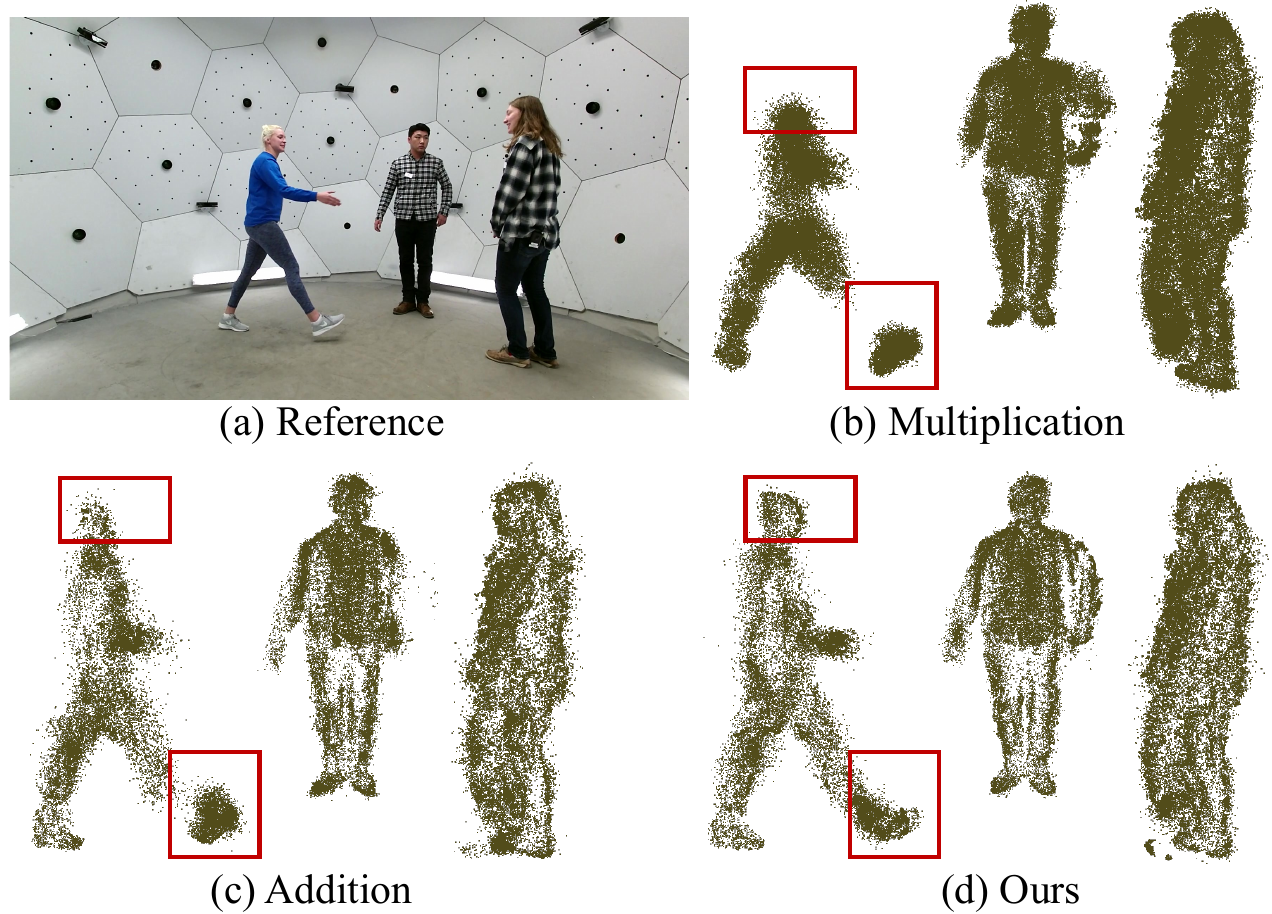}
 \vspace{-2mm}
 \caption{Visualization of learned Canonical Gaussian position distribution. Our opacity deformation approach learns more sparse and semantic distributions (hand and leg) in the canonical space than other opacity deformation approaches.}
 \vspace{-1em}
\label{fig:can_o_dist}
\end{figure}

\textbf{Effective Opacity Deformation.} In dynamic scenes with severe occlusion, it is crucial to model opacity changes effectively. We evaluate our opacity deformation technique against existing methods. Compared to the \textit{Addition}~\cite{gs4d} and \textit{Multiplication}~\cite{zhao2024gaussianprediction} methods, our \textit{Composition} maintains opacity in a binary state throughout training, as Figure~\ref{fig:train_opacity} shows. This results in a well-distributed and compact point representation in the canonical space, as illustrated in Figure~\ref{fig:can_o_dist}. In contrast, using other opacity deformation methods can lead to irrational shape changes and noisy distributions, resulting in poorer geometry, as shown in Figure~\ref{fig:ablate_od}.

\noindent\textbf{Foreground Mask Loss.} Previous regularization techniques from~\cite{2dgs} can cause elongated 2D Gaussians, particularly in the silhouettes of moving objects, due to inadequate separation of foreground and background. To address this, we use an alpha mask for supervision, which improves the geometry by providing clearer separation.

\begin{figure}[t]	\includegraphics[width=1.\linewidth]{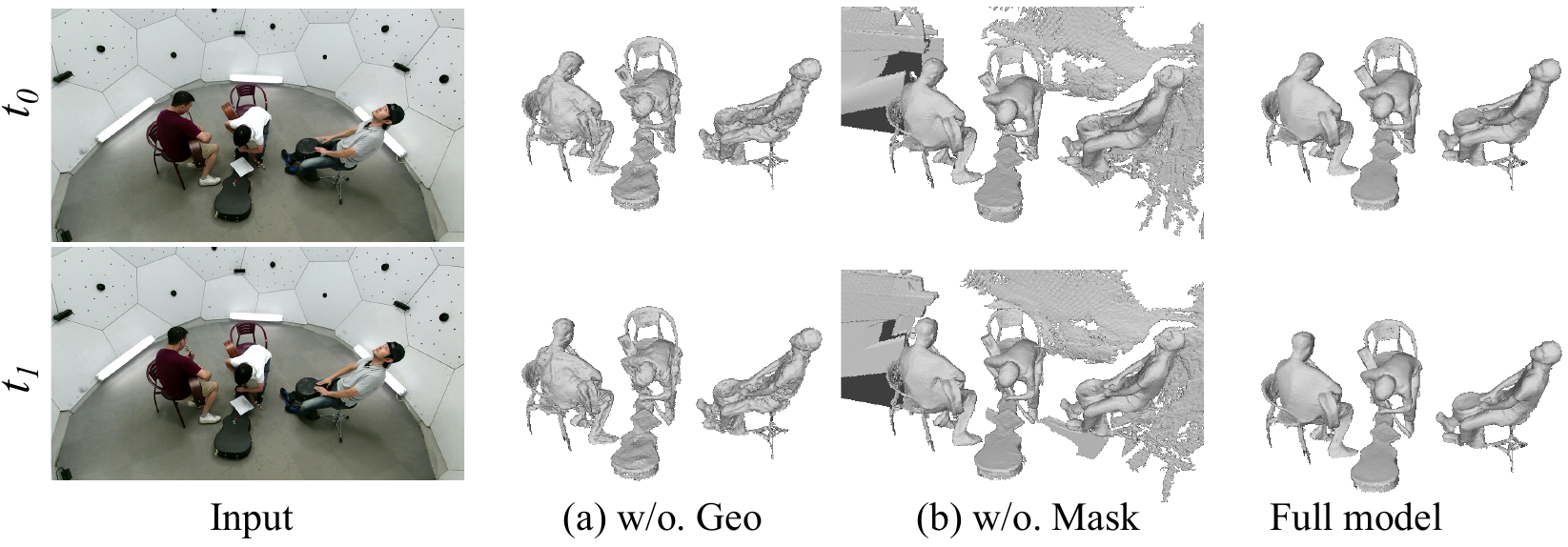}
 \vspace{-5mm}
 \caption{Qualitative studies for the regularization effects. 
Turning off the geometry regularization leads to a noisy surface; conversely, omitting the foreground mask results in a chaotic background. Our full model captures sharp and flat features.
}
\label{fig:ablate_full}
\end{figure}

\noindent\textbf{Geometry Regularization.} We also assess the impact of normal consistency and depth distortion regularization terms from 2DGS~\cite{2dgs} when applied to the deformed Gaussian space. Our complete model (Table~\ref{tab:ablation_full}) demonstrates the best performance. We find that omitting these regularization terms results in a noisy surface, as depicted in Figure~\ref{fig:ablate_full}.

\input{table/od}

\input{table/ablation}

%% file: table/dnerf_res_sub.tex
\begin{table}[t]
\vspace{-2pt}
\centering
\renewcommand{\arraystretch}{1.5}
\resizebox{0.48\textwidth}{!}{
    \begin{tabular}{ccccccccccccccccc}
    \toprule
    \multirow{2}{*}{Method} &  & \multicolumn{3}{c}{Lego$\dagger$} &  & \multicolumn{3}{c}{Mutant}        \\ \cline{3-5} \cline{7-9} 
               &  & $\text{PSNR}$ $\uparrow$ & SSIM $\uparrow$ & LPIPS $\downarrow$ &  & $\text{PSNR}$ $\uparrow$ & SSIM $\uparrow$ & LPIPS $\downarrow$ \\ \hline
    3DGS     &  &22.10&0.9384&0.0607&  &20.64&0.9297&0.0828&  \\
    TiNeuVox &  &26.64&0.9258&0.0877&  &30.87&0.9607&0.0474&  \\
    HexPlane   &  &\cellcolor{tabthird}26.67&\cellcolor{tabthird}0.9386&\cellcolor{tabthird}0.0544&  &\cellcolor{tabthird}33.67&\cellcolor{tabthird}0.9802&\cellcolor{tabthird}0.0261&  \\
    4DGS &  &\cellcolor{tabfirst}27.01&\cellcolor{tabsecond}0.9427&\cellcolor{tabsecond}0.0533&  &\cellcolor{tabfirst}37.59&\cellcolor{tabsecond}0.9880&\cellcolor{tabsecond}0.0167&  \\
    DG-Mesh &  &23.05&0.9014&0.1115&  &30.40&0.9680&0.0550&  \\
    Ours       &  &\cellcolor{tabsecond}26.77&\cellcolor{tabfirst}0.9467&\cellcolor{tabfirst}0.0508&  &\cellcolor{tabsecond}36.88&\cellcolor{tabfirst}0.9905&\cellcolor{tabfirst}0.0098&  \\
    \bottomrule
    \end{tabular}
    }
    \caption{We report PSNR, SSIM and LPIPS score for the D-NeRF dataset \texttt{Lego} and \texttt{Mutant} scenes. $\dagger$ indicates we follow the test set split method for the \texttt{Lego} scene from DeformableGS~\cite{deformable3dgs}.
 Our method achieves a comparable rendering quality. The \colorbox{tabfirst}{best},  \colorbox{tabsecond}{second best} and \colorbox{tabthird}{third best} are denoted by pink, orange and yellow, respectively.}
    \label{tab:main_res_dnerf}
\end{table}

%% file: table/training_time.tex
\begin{table}[h!]
\centering
\resizebox{0.95\linewidth}{!}{
\begin{tabular}{lcccccc}
\toprule
& \textbf{Accuracy $\downarrow$} & \textbf{Completion $\downarrow$} &
\textbf{Average $\downarrow$} &
\textbf{Time $\downarrow$} & 
\textbf{Real-Time} \\
\midrule
Tensor4D~\cite{shao2023tensor4d} & 16.1 & 25.2 & 21.2 & $\sim$14h & \ding{55} \\
4DGS~\cite{gs4d} & 11.5 & 19.9 & 15.7 & $\sim$1h & {\checkmark}  \\
SDFFlow~\cite{sdfflow} & 11.7 & 19.5 & 15.7 & $\sim$14d & \ding{55} \\
Ours & \textbf{10.1} & \textbf{19.2} & \textbf{14.6} & $\sim$1h  & {\checkmark}  \\
\bottomrule
\end{tabular}
}
\caption{Performance comparison on the CMU Panopic dataset~\cite{cmu}. We report the chamfer distance, training time and inference speed.}
\label{tab:training_time}
\end{table}

%% file: table/od.tex
\begin{table}[t]
\centering
\resizebox{0.5\textwidth}{!}{
\begin{tabular}{lcccc}
\toprule
& \textbf{Accuracy $\downarrow$} & \textbf{Completion $\downarrow$} & \textbf{Average $\downarrow$} &
\textbf{Points Number}  \\
\midrule
A. w/o OD & 10.7 & 19.8 & 15.2 & 241,232 \\
B. Multiplication & 10.8 & 19.8 & 15.3 & 235,281 \\
C. Addition & 10.8 & 19.6 & 15.2 & 95,951 \\
D. Composition (Ours) & \textbf{10.1} & \textbf{19.2} & \textbf{14.6} & 96,109 \\
\bottomrule
\end{tabular}
}
\caption{Quantitative studies for the different opacity deformation methods on the CMU Panoptic dataset.}
\label{tab:ablation_od}
\end{table}

%% file: table/ablation.tex
\begin{table}[t]
\centering
\resizebox{0.5\textwidth}{!}{
\begin{tabular}{lccc}
\toprule
& \textbf{Accuracy $\downarrow$} & \textbf{Completion $\downarrow$} & \textbf{Average $\downarrow$} \\
\midrule
A. w/o geometry regularization & 14.6 & 19.9 & 17.3 \\
B. w/o opacity deformation & 10.7 & 19.8 & 15.2 \\
\midrule
C. Full Model & \textbf{10.1} & \textbf{19.2} & \textbf{14.6} \\
\bottomrule
\end{tabular}
}
\caption{ Quantitative studies for the regularization terms on the CMU Panoptic dataset.}
\label{tab:ablation_full}
\vspace{-10pt}
\end{table}

%% file: sec/5_conclusions.tex
\section{Conclusion} 
\label{sec:conclusions}
\noindent
We propose Space-time 2D Gaussians Splatting, a method that enables high-quality reconstruction of complex dynamic scenes from sparse view videos or even monocular videos. Space-time 2D Gaussians Splatting jointly optimizes canonical 2D Gaussian and voxel-plane-based deformation fields through geometry and photo-metric optimization scheme capable of reconstruction of real-world complex dynamic scenes. Extensive experiments conducted on the CMU Panoptic dataset and the D-NeRF dataset demonstrate the effectiveness of our method.

%% file: sec/X_suppl.tex
\clearpage

\appendix
\section{Datasets and Implementation Details}
\label{sec:rationale}
\subsection{More Implementations Details}
\begin{figure}[h]
\includegraphics[width=1.0\linewidth]{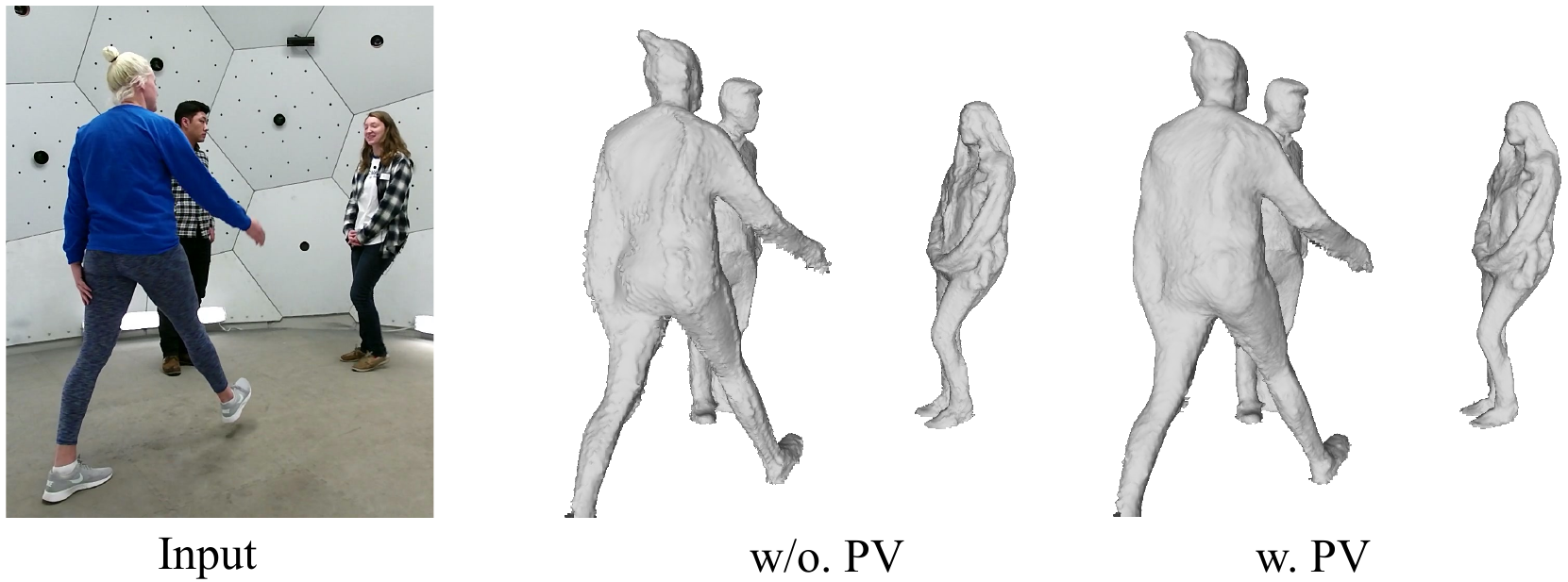}
 \caption{Qualitative ablation studies on mesh extraction with respect to the use of pseudo views. The incorporation of pseudo views leads to smoother mesh reconstruction.}
 \vspace*{-5pt}
 \label{fig:ablate_pv}
\end{figure}

\begin{figure}[h]
\includegraphics[width=1.0\linewidth]{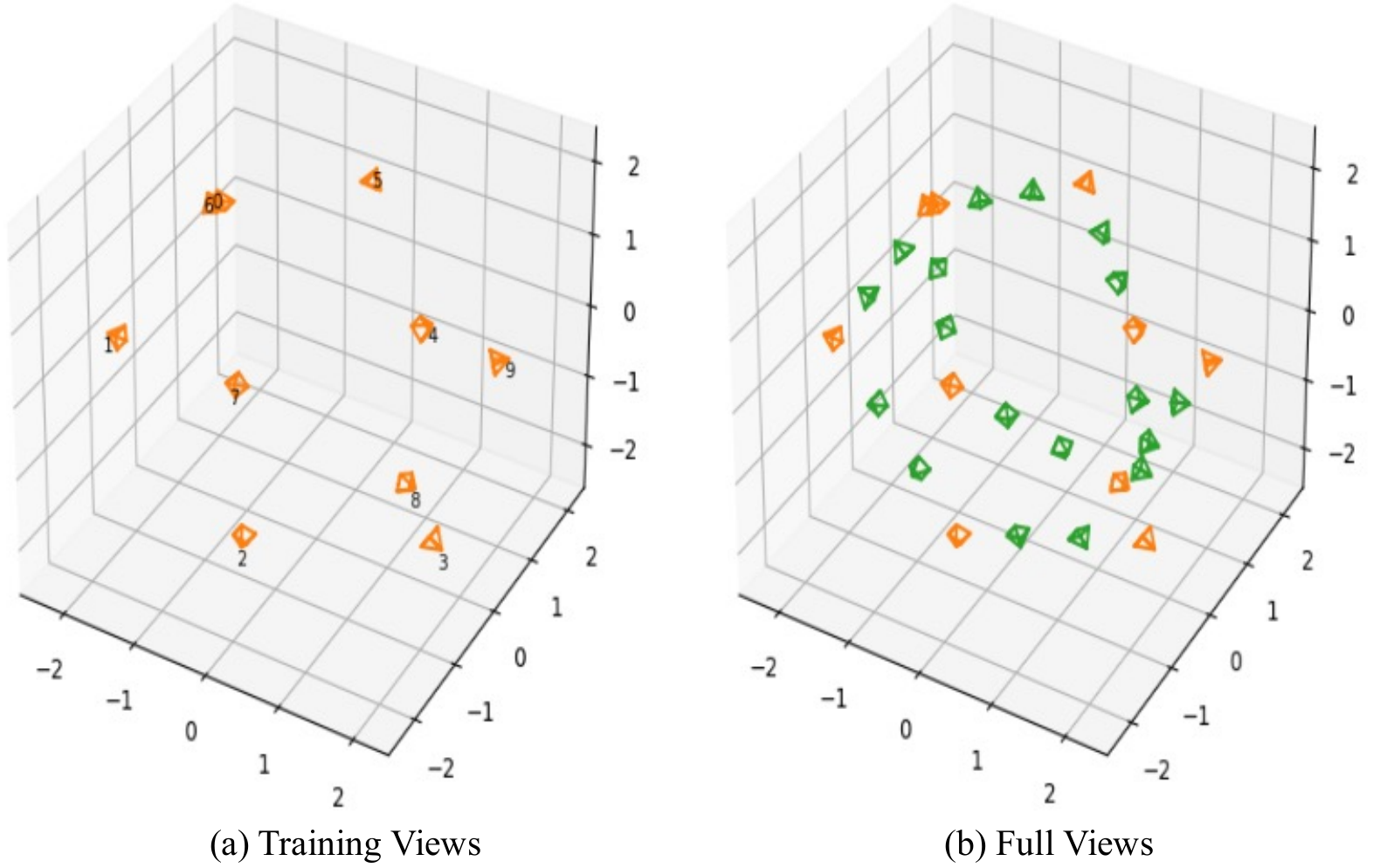}
 \caption{Visualization of the camera positions and orientations for the training views in (a) and the full views incorporating pseudo views in (b).}
 \vspace*{-5pt}
 \label{fig:cam_vis}
\end{figure}

\textbf{Mesh Extraction.} 
We additionally render pseudo views (PV) during TSDF Fusion~\cite{zhou2018open3d} to enhance sparse view mesh extraction. The training and pseudo views are illustrated in Figure~\ref{fig:cam_vis}\textcolor{red}{(b)}. However, using pseudo-views may result in over-smoothing or outliers. To balance mesh smoothness and geometric accuracy, we uniformly interpolate two views between each adjacent pair of training views, as depicted in Figure~\ref{fig:cam_vis}\textcolor{red}{(a)} (with indices indicating adjacency). The qualitative and quantitative comparisons between using and not using pseudo views are presented in Figure~\ref{fig:ablate_pv} and Table~\ref{tab:ablation_pv}, which shows that our approach slightly influences the accuracy but significantly reduces the noise of extracted mesh.

\begin{figure}[h]
\includegraphics[width=1.0\linewidth]{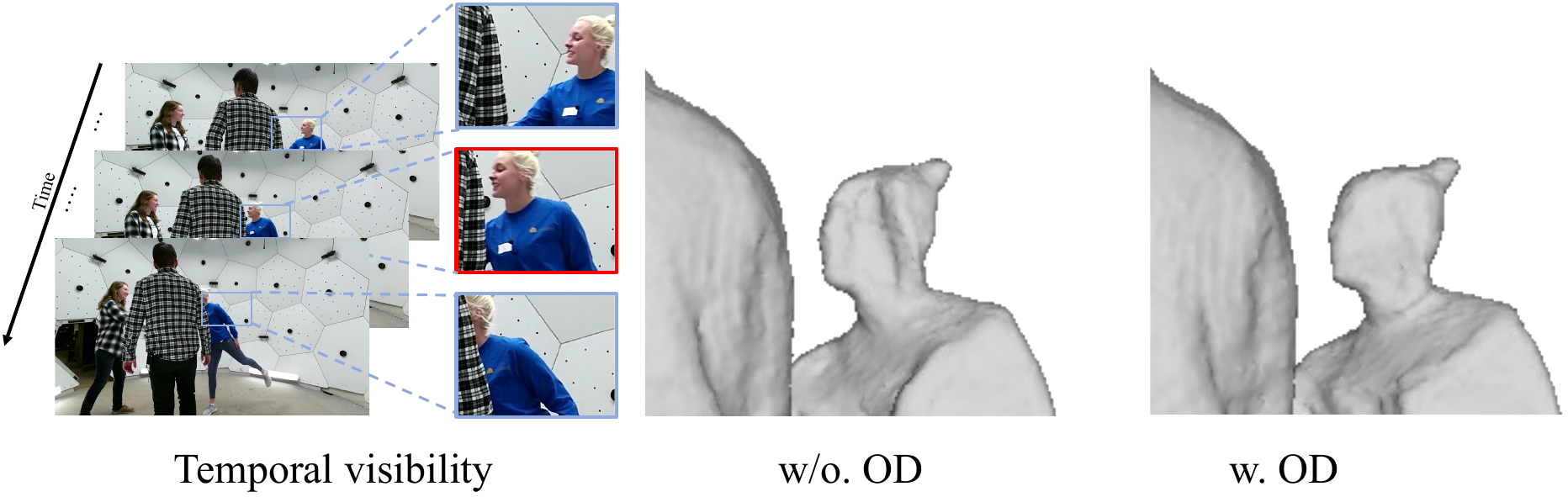}
 \caption{Qualitative ablation studies of opacity deformation on temporal visibility regions. Our method adapts to occlusions and reconstructs more accurate geometry.}
 \vspace*{-5pt}
 \label{fig:mesh_od_vis}
\end{figure}

\begin{figure}[h]
\includegraphics[width=1.0\linewidth]{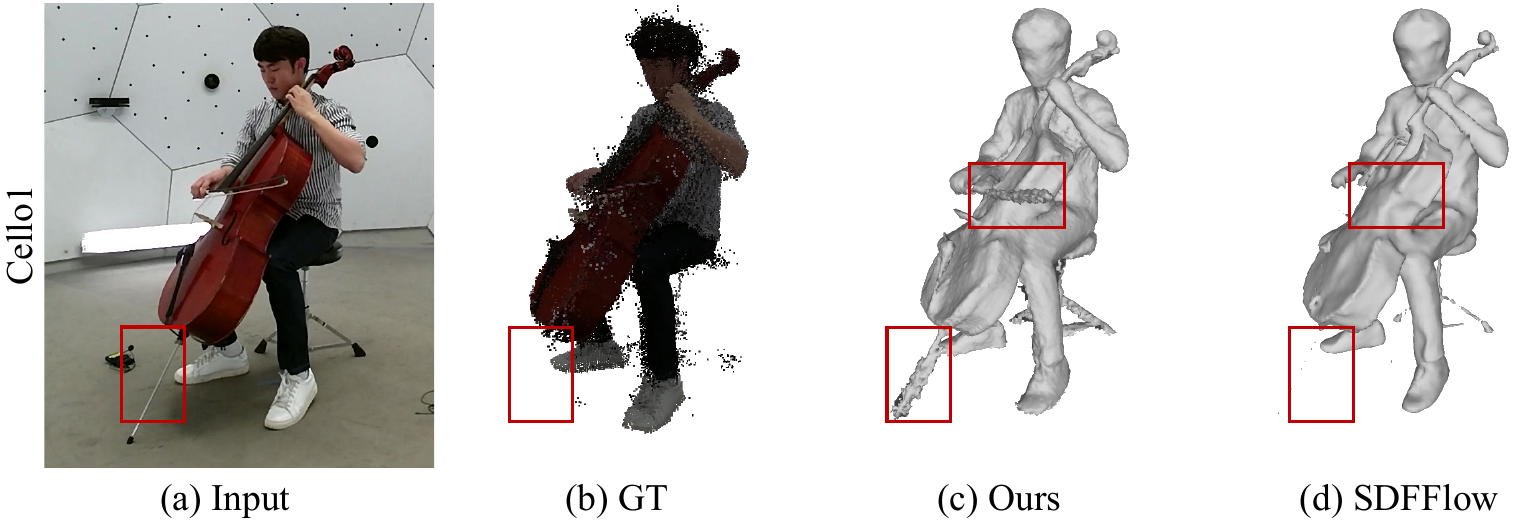}
 \caption{Compared with the input image (a), the ground truth (b) loses some components such as the bracket. Our method (c) can capture these structures compared to SDFFlow (d).}
 \vspace*{-5pt}
 \label{fig:cello}
\end{figure}

\begin{figure*}[t]	\includegraphics[width=1.\linewidth]{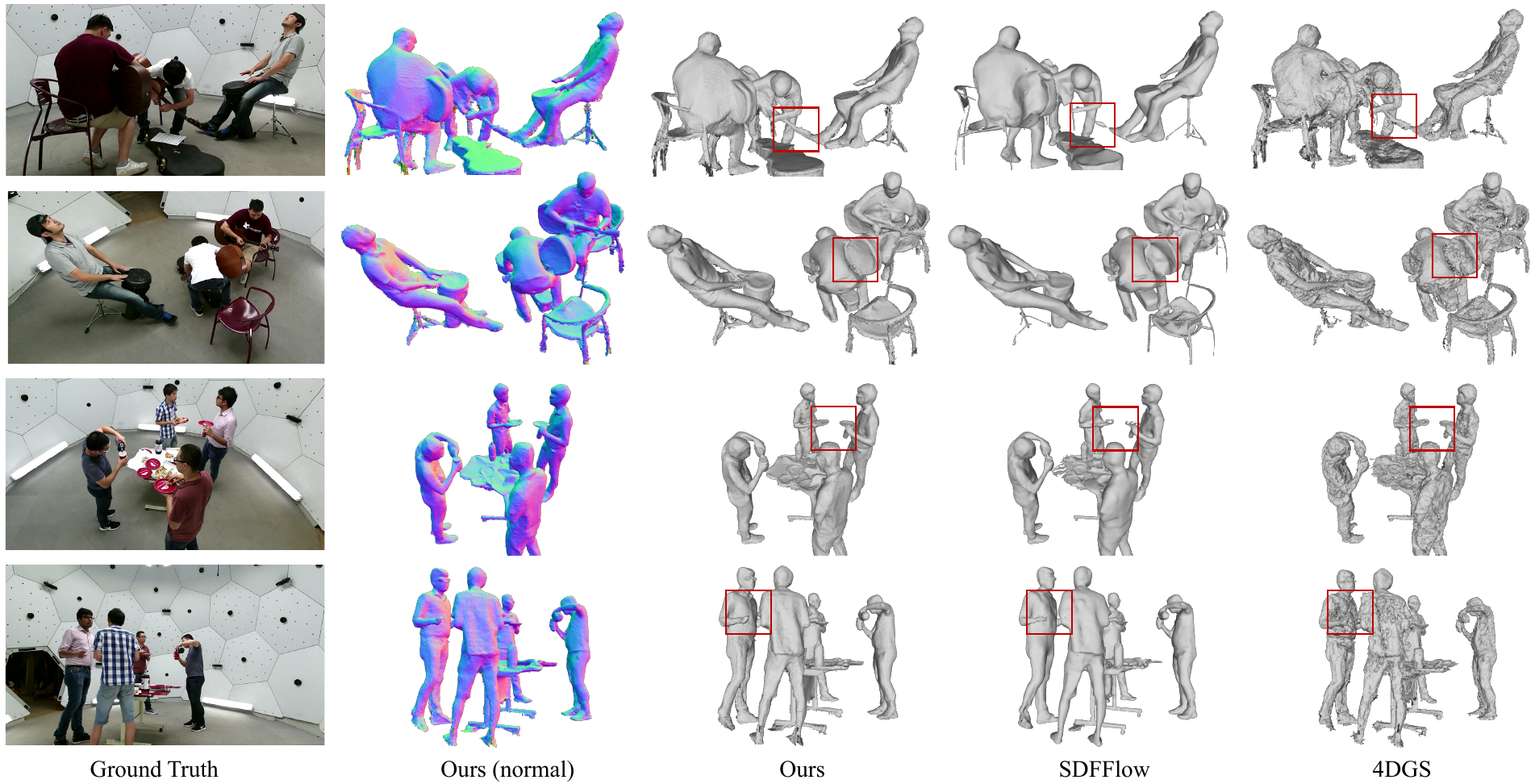}
 \caption{Visual comparisons at two different timestamps between our method, SDFFlow~\cite{sdfflow}, and 4DGS~\cite{gs4d} are conducted using scenes \texttt{Band1} and \texttt{Pizza1} from a real-world dataset~\cite{cmu}. }
 \vspace{-1em}
\label{fig:main3}
\end{figure*}

\subsection{Dataset Details}
Compared to the five scenes used by SDFFlow~\cite{sdfflow}, derived from CMU  Panoptic dataset~\cite{cmu}, we excluded \texttt{Cello1} due to inaccuracy in the ground truth. As shown in Figure~\ref{fig:cello}, compared with input observation, the ground truth loses some components such as the bracket. Although, our method can capture these structures as illustrated in Figure~\ref{fig:cello}\textcolor{red}{(b)}, quantitative performance drop compared to SDFFlow. We also observe the scale matrix of the \texttt{Band1} scene was inconsistent with the other scenes, so we adjusted it to approximately the same as other scenes. The proper scale matrix is crucial for Gaussian Splatting training, as it directly affects the learning rate of the Gaussian parameters.
\input{table/ablation_pv}
\section{Additional Results}
\label{sec:more_results}
We visualize the temporal visibility regions and opacity deformations in Figure~\ref{fig:mesh_od_vis}, which shows that our approach adapts to occlusions and results in accurate geometry.
Comparisons at two different timestamps of our method, SDFFlow~\cite{sdfflow}, and 4DGS~\cite{gs4d} are conducted using scenes \texttt{Band1} and \texttt{Pizza1} from the real-world dataset~\cite{cmu} are shown in Figure~\ref{fig:main3}. Our method, along with 4DGS, captures more details than the SDFFlow by leveraging the advantages of a point-based method. However, because 4DGS is an extension of 3DGS, it suffers from insufficient geometric accuracy and produces a noisy surface. We provide video clips showcasing the reconstruction results on the CMU Panoptic dataset in the supplementary materials. These videos concatenate the training view inputs, SDFFlow results, and our method for the subsequent image sequences, which have been compiled into videos at a frame rate of 5 FPS. Due to the long training time required for SDFFlow, some sub-scenes are hard to converge with our limited computational resources.


%% file: table/ablation_pv.tex
\begin{table}[t]
\centering
\resizebox{0.48\textwidth}{!}{
\begin{tabular}{lccc}
\toprule
& Accuracy $\downarrow$ & Completion $\downarrow$ & Average $\downarrow$ \\
\midrule
w/o PV & \textbf{9.7} & \textbf{19.1} & \textbf{14.4} \\
w PV & 10.1 & 19.2 & 14.6 \\
\bottomrule
\end{tabular}
}
\caption{Quantitative studies on the effects of using pseudo views for mesh extraction on the CMU Panoptic dataset. Although the use of pseudo views results in a slight decrease in geometric accuracy, it produces smoother reconstruction results.}
\label{tab:ablation_pv}
\vspace{-10pt}
\end{table}